\documentclass[10pt, a4paper]{article}
\usepackage{verbatim}
\usepackage[final]{lrec2026} 

\title{\textsc{TestiMole-Conversational}: A 30-Billion-Word Italian Discussion Board Corpus (1996--2024) for Language Modeling and Sociolinguistic Research}

\name{Matteo Rinaldi*, Rossella Varvara*, Viviana Patti*} 

\address{*Dipartimento di Informatica, University of Turin, Italy \\
         *Corso Svizzera 185, 10149 Torino, Italy\\
         matteo.rinaldi@unito.it, rossella.varvara@unito.it, viviana.patti@unito.it\\
         }

\abstract{
We present \textsc{TestiMole-Conversational} a massive collection of discussion boards messages in the Italian language. The large size of the corpus, almost 30B word-tokens (1996–2024), brings challenges in the processing and curation of the resource, but it renders it an ideal dataset for native Italian Large Language Models'  pre-training. Furthermore, discussion boards' messages are a relevant resource for linguistic as well as sociological analysis.
The corpus captures a rich variety of computer-mediated communication, offering insights into informal written Italian, discourse dynamics, and online social interaction in a wide time span. Beyond its relevance for NLP applications such as language modelling, domain adaptation, and conversational analysis, it also support investigations of language variation and social phenomena in digital communication.
\\ \newline  }

\begin{document}

\maketitleabstract

\section{Introduction}

Over the past three decades, a new form of written communication has emerged due to the diffusion of digital communication networks among the general public. This constituted a revolutionary event in the history of written language, as the 
digital medium began to be massively used as a form of communication meant for ordinary, spontaneous and interpersonal conversations, a domain that was previously addressed mainly by the oral form of communication. Researchers have coined different terms to refer to this variety, e.g., {\em computer-mediated communication} (CMC), {\em netspeak}, {\em online communication} (for the Italian language, \textit{e-taliano}, \textit{italiano digitato} or \textit{italiano neomediale}, \citealp{pistolesi18}).
This type of language is at the crossroad between oral and written language: despite the written medium, it has many features of informal oral communications; moreover, it has became a pervasive form of written communication even for less scholarized people, a social group that previously had orality as a primary form of communication 
 \cite{antonelli2016taliano}.

However, although there may be common features to all instances of CMC\footnote{We will use this term as a hypernym for denoting every type of language written with digital tools, not only computers, but also smartphones and tablets.}, different varieties of this type of communication can be recognized. For instance, the language used by a journalist in a blog post will be different from text messages sent through an instant messaging app among family members. In this landscape, a specific type of communication can be identified in the web-based forums and newsgroups. Among different platforms users have been able to communicate in {\em discussion boards} in asynchronous way, exchanging ideas on specific topics.


In this work, we present a corpus of messages exchanged within the online Italian-speaking community, gathering data from two technologies used by the general public for online discussion: Usenet’s newsgroups and several independently hosted forums\footnote{Unless otherwise specified, the cover term “discussion boards” will be employed to refer to both technological configurations, considered their similarity for the purposes of our work.}. The corpus comprehends 470 millions messages among forums and 90 millions Usenet, for a total of 30 billions word tokens.

The purpose of gathering such a large amount of this specific type of data is threefold. First, the creation of this corpus allows the data-driven linguistic analysis of the specific language variety of discussion boards, a dialogical type of CMC for which large scale studies are lacking in Italian. The diachronic nature of the corpus, where each post is annotated with the exact time of writing, provides the possibility to perform precise analysis regarding the evolution and change of the online language across time, both at the lexical and grammatical levels. It also gives the opportunity to capture specific orthographic forms that emerged and then declined in digital communication, such as SMS-style abbreviations or specific emoticon's styles.

Second, online resources are fragile and susceptible to the passing of time: 
resources needed to keep the pages online, such as servers or web-hosting, may cease to be maintained. Even if it is hard or impossible to give an estimate about how many online forums have existed since the advent of the web and are now no longer reachable, it is reasonable to expect that a large number of discussion boards are nowadays forever lost, together with all the messages exchanged by their users. This is the reason why archiving this material is a way to ensure its persistence throughout time, avoiding the risk of losing such a peculiar linguistic resource, as well as the large amount of information contained in the posts. 

Lastly, given the large size of the collected resource, it may constitute a viable option also as a pre-training dataset for Large Language Models (LLMs), whose training needs impressive amounts of data.
%
The current techniques for training large language models, whether for generative or classification purposes, require a massive amount of data to reach an acceptable level of performance. This poses concern when the goal is to train language models capable of correctly serving the needs of language communities other than English. In the case of Italian, for example, web-derived data that could be used as training data are several orders of magnitude scarcer than the material available for English\footnote{For example, in the Common Crawl dataset Italian accounts only for 2\% of the data, in comparison of 44\% of English. See \url{https://commoncrawl.github.io/cc-crawl-statistics/plots/languages.html}}.
A notable exception is the TWITA social media collection 
\cite{TWITAbasile-etal-2018-long}, a ten-year collection of Twitter posts in Italian (2012--2023), that has been exploited for the learning phase of AlBERTo, a BERT language understanding model for the Italian language \cite{alberto19}.
Discussion boards’ messages are an invaluable resource for training large language models: they provide large amount of written texts, originally written in Italian, usually in a direct, informal, and dialogical style that can be useful for developing systems designed to carry out conversations with users, such as chatbots. 
Moreover, their messages can be considered as  "information goldmine", especially when the topics concern highly technical information that is often hard to find in other resources. This information can be employed in LLM training in order to improve their knowledge on more niche themes. 
Exposing models to discussion boards' messages may improve problem-solving capabilities, given that a significant portion of the messages exchanged on discussion boards involves users asking for and receiving help to solve practical issues. 
Moreover, the concept of moderation and the way in which users and moderators handle heated discussion can provide the model with hints of how to correctly recognize and handle potential offensive, provocative, or belligerent tones.  For these reasons, we believe that a resource such as \textsc{TestiMole-Conversational}
 can add a significant value to the dataset used to train models for the Italian language.

In this paper, after a review of the related literature, we describe the corpus collection procedure, designed to retrieve a large amount of clean data.We present some of the challenges in corpus creation, such as clean extraction of data and their anonymization, as well as corpus statistics.

The JSONL files of the dataset are currently available at the following address \url{https://huggingface.co/datasets/mrinaldi/TestiMole}.

\section{Related Work}
The potentiality of gathering large text corpora from discussion boards was already explored more than thirty years ago. \cite{lund_producing_1996, burgess98}, notably, compiled the HAL Corpus by collecting 131 million words from Usenet over the course of February 1995. The HAL Corpus was used to train a model encoding semantic and grammatical meaning by transducing lexical co-occurrence. At the time, Usenet was valued as a reliable source for building cognitive models, owing to its conversational nature  and the breadth of topics it covered. The HAL corpus, later extended to 320 million words, was also used  to create a large database of proper names  \cite{conleyBH99}.
Corpora derived from Usenet were also used in at least two German language projects: the ELWIS corpus, compiled between 1992 and 1993, and employed to investigate  language use in newsgroups \cite{feldweg95};  the DeReKo project, the German Reference Corpus \cite{SchrockLungen2015}, into which an annotated version of the ELWIS corpus was integrated.

Other examples for the English language include the reduced redundancy Usenet Corpus \citelanguageresource{usenet13}, collected from 2005 to 2011 and containing more than seven billion words, the ``Usenet as a text corpus'', which comprehends 53,245 articles \cite{MahoneyUsenet2000}, and the corpus collected by \citet{Hoffmann22}, constituted by 773,772 messages from Usenet newsgroups.

As for Italian, the group coordinated by Manuel Barbera at the University of Turin developed the NUNC corpora, the largest collection of material drawn from  Italian Usenet prior to the current work  (127M tokens, \citealp{NUNC2}). Most notably, NUNC corpora are annotated and also available for other languages \cite{NUNC11} such as Spanish and French  
\citelanguageresource{NUNC}.

General web corpora contain, among others, texts from forum discussions as well, but not all resources provide metadata that allow the users to identify forum discussions. The most widely known web corpora include, but are not limited to, the WaCky corpora \cite{baroni2009wacky}, the COW corpora \cite{schafer2012building}, the SketchEngine's TenTen family \cite{jakubivcek2013tenten}, and the Aranea corpora \cite{benko2014aranea}. It is also worth mentioning that the collection of web corpora gained huge attention starting from the late 2000s, as shown by the intense research activity of the ACL special interest group on the Web as Corpus (SIG-WaC).

To date, \textsc{TestiMole-Conversational} is the largest Italian corpus of Usenet and Forum discussions; it can be successfully used for the development and improvement of NLP tools for the Italian language, as well as enabling unprecedented analysis of the language adopted by Italian speakers in the Web over almost thirty years.

Discussion board conversations have also been used to support sociological and psychological analyses of online communication behaviour. One example is the phenomenon of ``trolling'', whereby users engage in explicit or covert aggression toward others with the deliberate aim of provoking an emotional response in the victim \cite{Hardaker2010}.

The selective and topic-specific nature of online discussion boards, particularly forums, makes these platforms fertile grounds for marginalized and extremist groups alike. A notable example is the Italian ``Forum dei brutti" (`Forum of the ugly'), which serves as the primary online community for the Italians ``Incel'' subculture. This forum has been extensively studied due to its pronounced misogynistic stance, examined both from sociological perspectives \cite{cava_pasciuto_brutti23} and within the context of hate speech detection research in computational linguistics \cite{gajo-etal-2023-hate,gemelli-minnema-2024-manosphrames}.


\section{The \textsc{TestiMole-Conversational} Resource}

\subsection{Discussion Boards}
The text sources of the \textsc{TestiMole-Conversational} corpus are two types of discussion boards. Discussion boards are platforms where users can exchange messages on specific topics. Messages, called “posts”, are organized in “threads” that refer to a very specific topic of discussion, usually identified by the title of the discussion. Compared with instant messaging platforms, such as IRC (Internal Relay Chat), discussion boards are designed for asynchronous communication, where the same discussion can be continued for an indefinite amount of time. The topics allowed on a specific discussion board depend on the rules and characteristics of the given board: posts are organized in specific arbor-like hierarchies limiting the scope of the topics that are appropriate to write in a specific board; forums, on the other hand, are dependent on the decisions of the administrators so, depending on the platform, they can be more or less specific about the topics that intend to represent. In general, it is possible to open a limited number of discussions related to topics different from the  scope of the board, and this situation is indicated by the “off-topic” label, often shortened as “OT”.
Forums are usually moderated, that is, a restricted group of people is given the power to terminate discussions that are leading to personal attacks, as well as to invite users to avoid exacerbating tones. Moderators can also issue temporary or permanent bans to users who violate forum rules. Cursing, especially on larger and more important forums, is usually forbidden. Note, however, that moderation on Usenet was not generally applied, and  texts sent by users were not subject to any kind of censorship or restriction.

Compared to social networks, discussion boards are more focused on restricted topics. Such texts are often highly technical, providing a valuable source of hard-to-find information derived from users personal or professional experience.
Discussion board messages, indeed, contain detailed and very specific information written by people who are often passionate and very knowledgeable about what they are discussing. Nonetheless, 
alongside  truthful and useful data, users may share blatant misinformation, personal opinions, or errors, with or without being aware of it.

\subsubsection{Usenet}
Usenet was the first widely used platform for discussion boards on the Internet. Its inception dates back to 1979, when the Network News Transport Protocol (NNTP) enabled the exchange of messages between servers following a hierarchical organization of newsgroups. Articles propagate across servers by creating copies, with distribution managed through relationships between nodes. Unlike centralized forums hosted on single servers, Usenet operates as a distributed system where independent servers exchange articles through bilateral agreements, with no central authority controlling content distribution beyond community established procedures for creating new groups.
The diffusion of Usenet in Italy\footnote{The material to reconstruct the history of Italian Usenet was obtained from the official pages of the GCN-IT \url{https://www.news.nic.it/} as well as the personal website of one of the Italian Usenet's founders, Maurizio Codogno: \url{https://xmau.com/usenet/}} started significantly later than in other countries. While isolated discussion groups existed earlier, the structured Italian hierarchy \textit{it.*} was formally established in 1994-95 through a coordinated effort led by Alessio F. Bragadini and Stefano Suin in the context of the University of Pisa's SerRA Project. The initiative was formalized at the NIR-IT-2 conference in Milan on December 13, 1994, with the goal of creating a national discussion space independent from single institution or provider. Prior to this, the only Italian-language group was \textit{soc.culture.italian}, which mixed international discussions about Italy with domestic content\footnote{An earlier hierarchy ita.* had been created around 1993 by Marco Negri at the University of Milan's Department of Computer Science but failed due to poor coordination between Italian news servers and limited propagation outside the academic network.}. The new it.* hierarchy officially began operations in January 1995, initially comprising groups like \textit{it.politica}, \textit{it.sport}, \textit{it.spettacolo}, \textit{it.scienza}, and \textit{it.cultura}, designed to reflect traditional media organization. To generate initial traffic, bidirectional gateways were established with major Italian mailing lists, allowing the same content to be accessed both via newsreaders and list subscriptions. International distribution began in March 1995 and by November 1995, the hierarchy had achieved full integration into the worldwide Usenet system.
The creation of new newsgroups followed international Usenet standards, requiring 50 interested users or connection to mailing lists with 150 subscribers.
The \textit{Gruppo di Coordinamento NEWS-IT} (GCN) was established as a volunteer working group to manage the hierarchy, the \textit{Request For Discussion} (RFD) and \textit{Call For Votes} (CFV) procedures, and user documentation. The GCN decided to establish second-level sub-hierarchies (like \textit{it.comp}, \textit{it.cultura}, \textit{it.hobby}) to thematically organize groups, and to create \textit{it.binari.*} for binary files with mandatory moderation to prevent abuse. They also introduced several innovations: groups could be removed if traffic fell below 100 articles in three months, crossposting was limited to a maximum of 10 newsgroups, and binary file attachments were prohibited outside designated groups. Moderation emerged as a critical tool, with both human moderators and experimental ``robomoderation'' systems that automatically filtered messages based on technical criteria like crosspost count or binary content. Netiquette rules, sometimes referred to in Italian as ``galareteo'' or ``retichetta'', were actively promoted through FAQ files\footnote{\url{https://www.news.nic.it/doc/emily.html}}. The GCN faced recurring challenges regarding content control, particularly concerning illegal material and copyright infringement, leading to policy discussions with GARR authorities and commercial providers about legal responsibilities. Each newsgroup required a "manifesto" (\textit{charter}) defining its scope and acceptable topics. 
Newsgroups in Italy were the main form of discussion used in Italy between the late 1990s and the beginning of 00s, alongside with Bullettin Board System (BBS), which were also accessible via a direct dial-up connection. Forums, hosted on private platforms, began to supersede Usenet around 2004,
according to our statistics (see fig.1).
\begin{figure}
    \includegraphics[width=0.48\textwidth]{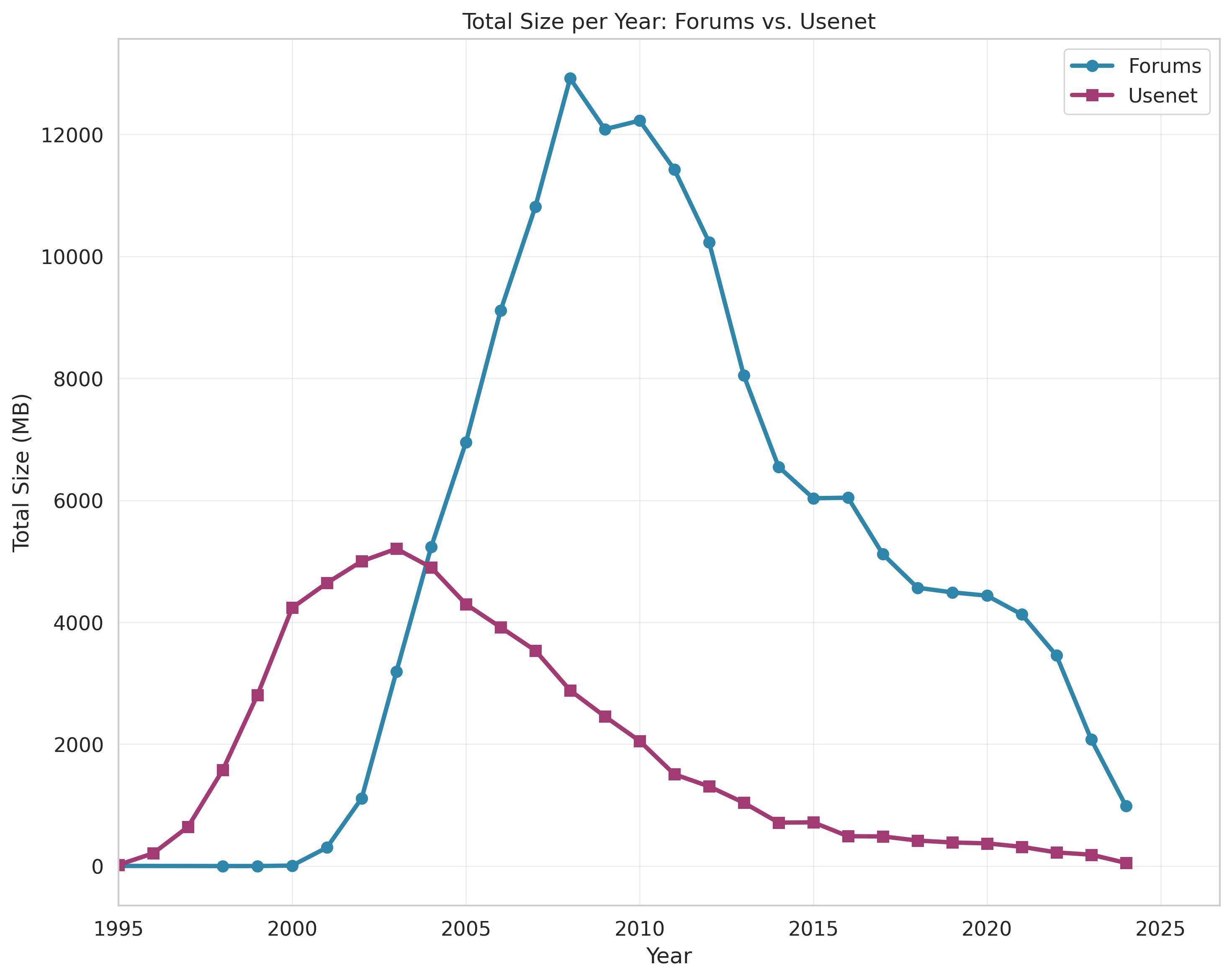}
    \caption{Total corpus size (in MB) per year. Forum overtakes Usenet around 2004.}
    \label{fig:temporal}
\end{figure}

\subsubsection{Forums}
Forums represent an evolution of online discussion spaces after Usenet, offering a more centralized environment for asynchronous communication. Unlike the distributed architecture of Usenet, forums are hosted on dedicated web servers managed by administrators who define both thematic organization and access policies. Each forum is typically structured hierarchically, with sections and subsections grouping discussions according to topic, within which users can post individual messages. Registration is generally required to participate, allowing the system to manage user identities, moderation privileges, and content visibility. Forums usually rely on software frameworks, such as vBullettin, phpBB or Simple Machine Forum, accessible via web-browser instead of a dedicated newsreader software like Usenet. Moderation in forums is usually more formalized than in Usenet: administrators and moderators enforce rules of conduct, prevent abusive or off-topic content, and ensure the overall coherence of discussions. The introduction of moderation tools such as warnings, temporary bans, and message deletion contributed to the development of online communities characterized by hierarchies and social norms.

With their diffusion in the late 1990s and early 2000s, forums became one of the main hubs of online interactions, often forming communities about highly specific interests. Thematic specialization, indeed, was one of the main traits of forums: distinct platforms emerged for every conceivable topic, from technical discussions to entertainment, forming a form of collective knowledge production. Users' reputations were built solely through the accumulation of contributions, rather than through algorithmic amplification based on engagement metrics, as is the case with current social media. The forum model thus privileged temporal continuity and participation history over immediate visibility. As a form of meritocratic social hierarchy, seniority and posting frequency determined the level of recognition within the community.

A distinctive aspect of forums was their semiotic dimension: users developed and shared linguistic conventions characterized by abbreviations, neologisms, and  acronyms (such as "trolling","flame","newbie"/"niubbo").

Despite their subsequent decline with the advent of social networks, forums have persisted as a form of online communication. Their decentralized and topic-oriented architecture constitutes a different paradigm compared to the algorithmically driven and feed-based structure of social networks: forums tend to promote discussions focused on specific contents, while social networks privilege transient engagement.

\subsection{Collection Methodology}

The material was obtained through a web scraping effort going on from February 2024 to May 2024. 
The texts more distant in time date back to 1996.
To create the data set, several scripts were developed using Python3 and libraries such as BeautifulSoup and Selenium. The scripts were often manually crafted for each resource in a very slow but precise process, even if the overall structure of the code was generally the same.
In particular, all the scraped forums were based on a limited number of discussion board platforms, such as vBulletin, phpBB, XenForo, Simple Machines Forum, Invision, and Snitz. Often, it was possible to use the same script designed for a forum based on a specific platform (e.g., phpBB) for a different forum sharing the same platform. However, this was not always the case: forum software usually allows for a great degree of customisation, through administrators' settings or the adoption of plugins, and such differences required different parsing patterns to be configured in each scraping script.
Thanks to the flexibility of the BeautifulSoup library, most of the work was limited to identifying the correct HTML identifiers, a task made easier by modern browsers equipped with developer consoles such as Mozilla Firefox. The BeautifulSoup library provides the programmer with helpful \textit{syntactic sugar} that can be used, for example, to find all the elements sharing a specific class
, iterate over them
, perform additional operations on the identified objects
, and so on. 

In general, the first step was to identify the URL pattern logic for each topic published in the forum. For example, if the forum discussions have a URL such as \texttt{FORUM \_BASE\_URL/viewtopic.php?t=N}, where \texttt{N} uniquely identifies the discussion, it is then possible to cycle through all the discussions from number 0 up to the latest one, which is manually identified by checking the most recently posted message at the time of scraping (although this verification could probably be automated). In some cases, a SEO-friendly string is appended to the URL, for example, \texttt{FORUM\_BASE\_URL/category-name/N/title-of-the-discussion}. 

After having identified the URL pattern, the next step was to understand the logic that the discussion board used to handle threads with many messages: all discussion board platforms include a pagination mechanism in order to avoid loading a large number of messages all at once. In general, it was possible to identify a specific HTML identifier for the container of pagination links (e.g., \verb|<div class="pagination">|), and to retrieve the last link in the pagination container.
Pagination was one of the trickiest parts of the scraping process, due to the high variability we found across the scraped forums. In some cases, it was possible to obtain a more regular URL schema by using the ``printer-friendly'' version of the page, when available.

The remaining useful information in the web pages was generally easier to identify. For each discussion, we were interested in retrieving all the messages, and finding the correct rule was straightforward, because most platforms use a unique identifier for each post (e.g., \verb|<div class="post">|, \verb|<li class="message">|, \ldots), making it possible to iterate over the portions of HTML code pertaining to each post. After having identified the logic for isolating individual posts, it was necessary to locate the identifiers for the required fields, namely the author of the post, the time and date of posting, and the body of the message. Again, manual inspection was used to identify the HTML identifier for each field, although these identifiers were often shared across different forums. Parsing the date was trickier: in the most straightforward cases, a special HTML \verb|<datetime>| tag was present, making it trivial to extract the message timestamp. Conversely, when only the written Italian form of time and date was present (e.g., \textit{Alle 16.30 di Domenica, 14 Dicembre 1997}), more effort was required to reconstruct the machine-readable format of the date (in this case, \texttt{1997-12-14T16:30:00}).

Although all the scripts were manually edited and refined for each forum, it is possible to create a ready-to-use pipeline compatible with at least all the forums already scraped, and potentially many more. This has not yet been accomplished; however, 
we plan to release a tool ready to use that could be leveraged for other research purposes. Further details about the scripts are reported in Appendix.

From each URL, each post was scraped and added to the dataset as a single row, containing the following metadata:
\begin{itemize}
\setlength{\itemsep}{0pt}\setlength{\parskip}{0pt}
    \item “title”: //the title of the thread;
	\item “author”: //anonymized ID of the post’s author;
\item	“id”: //unique identifier of the thread in the groups’ threads;
\item	“progressive\_id”: //identifier of the post in the thread; counter starts from 1;
\item	“timestamp”: //the time and data of creation of the post, in ISO-8601 format;
\item	“newsgroup”/”forum”: // the identifier of the specific newsgroup or forum;
\item	“text”: // the body of the message.
    \end{itemize}

In order to provide a pseudonymization of the dataset, each value present in the field "author" was substituted with a progressive number. In this way, the relationship between the authors and the posts was kept, but the specific usernames were hidden.

The computational resources required to run the scraping scripts were minimal: an old dual-core laptop from 2006 was sufficient, and its processing power was not a bottleneck relative to the network speed (domestic ADSL).

The result is a clean dataset that can be used with few 
preprocessing because the extraction scripts retrieved text data directly in a structured way, without the need to apply further filtering.

While the list of Italian Usenet hierarchies is straightforward to obtain from official sources, for forums we used a search engine to look for potential candidates: the forums selected for scraping were identified using keywords such as \texttt{"forum"}, \texttt{"viewtopic"}, and \texttt{"showthread"}. No strict criteria were adopted for the selection of forums. However, larger forums were preferred, while forums with very few messages (<1,000) or containing only spam were discarded.


\subsection{Corpus Statistics}

Overall, the \textsc{TestiMole-Conversational} corpus contains almost 30 billions word-tokens\footnote{The exact size of the corpus is 29,592,255,016 tokens.}, 
with a larger part devoted to forum texts (23 billions vs 7 billions for Usenet data). 

\begin{figure}
    \centering
    \includegraphics[width=\linewidth]{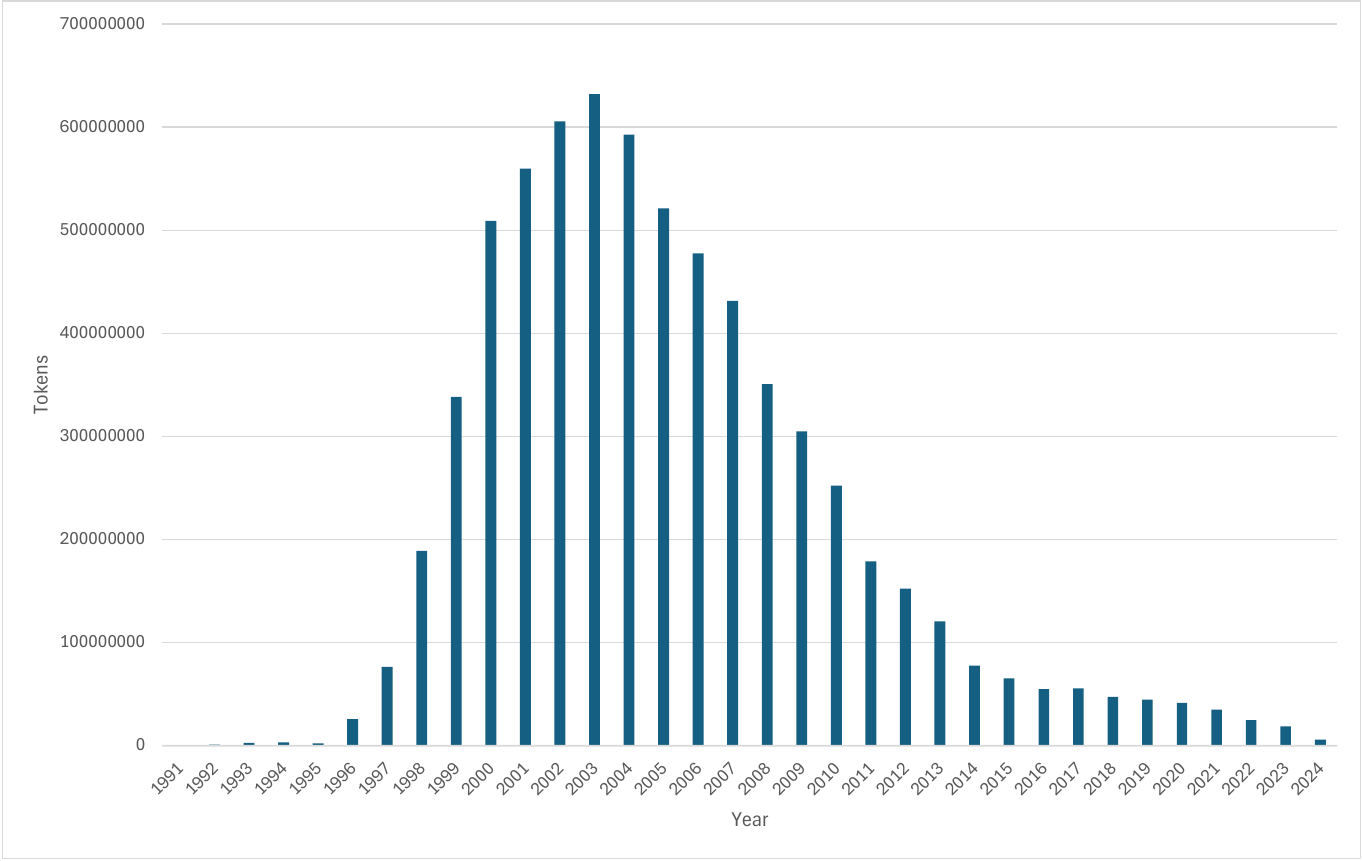}
    \caption{Usenet - Number of tokens per year}
    \label{fig:ntokens_usenet}
\end{figure}

\begin{figure}
    \centering
    \includegraphics[width=\linewidth]{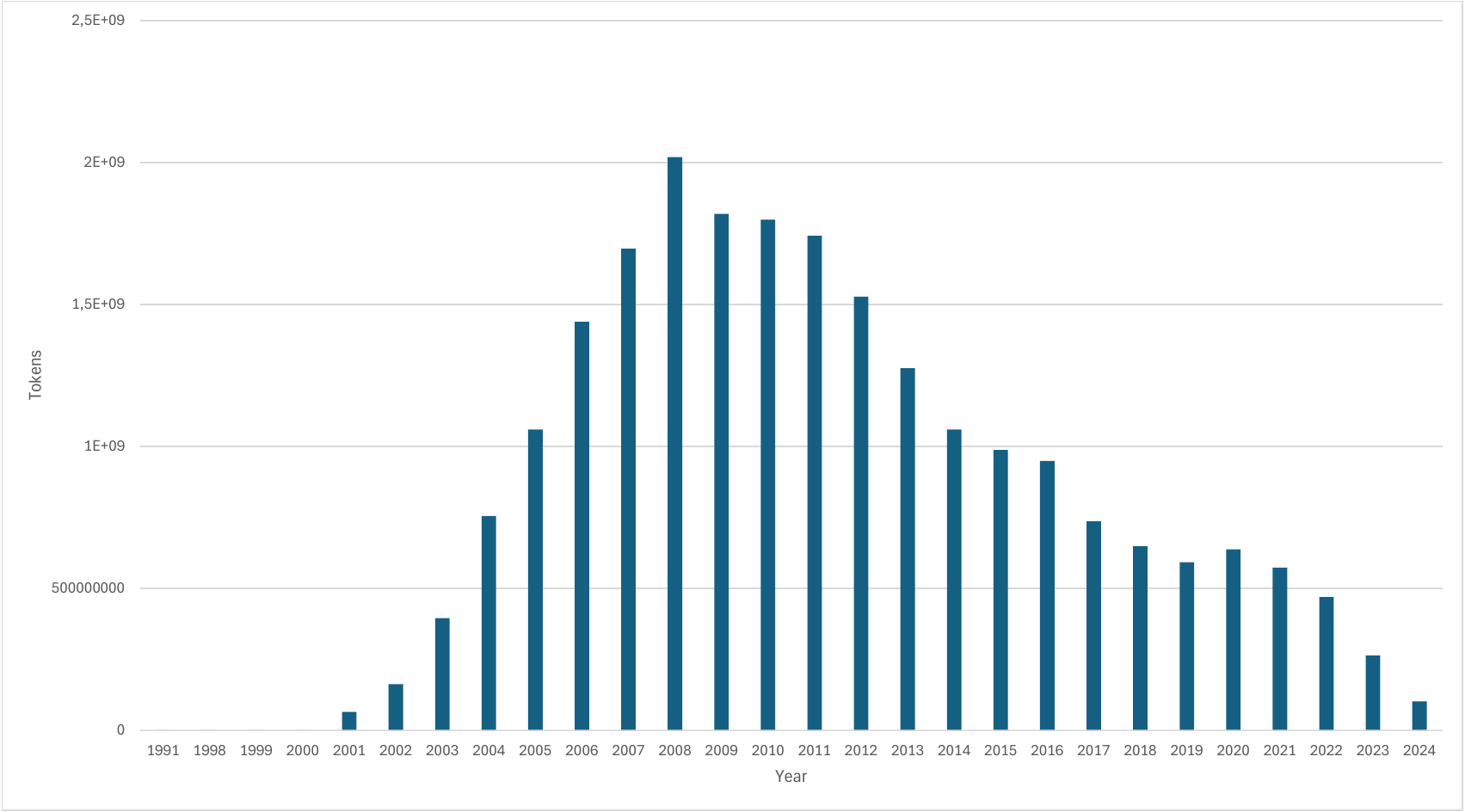}
    \caption{Forums - Number of tokens per year}
    \label{fig:ntokens_forums}
\end{figure}

Larger amount of data were collected for the central years of the time span considered (between 2003 and 2011, see fig. \ref{fig:ntokens_usenet} and \ref{fig:ntokens_forums}): pages from the nineties were more rare, probably because no longer maintained. In more recent years, instead, this type of discussion boards has probably lost success, with other platforms (such as social networks) becoming digital places for discussions. It is interesting to note that the distribution of tokens among years differ for the two subgroups of data: Usenet data reach the peak of tokens in 2003, while for forums data we have the highest number of tokens for the year 2008. It seems indeed that Usenet's popularity decreased earlier than those of other forums (see fig.\ref{fig:temporal}).

In terms of number of posts, the Forum corpus contains 468,391,746 posts in 25,280,745 unique threads (on average, 18.5 posts per thread), while the Usenet corpus consists of 89,499,446 posts in 14,521,548 threads (on average, 6 posts per thread). The computation of tokens with a sub-word tokenizer employed for LLM training (Tiktoken BPE tokenizer, model \textit{cl100k\_base}) resulted in 20B and 62B tokens for Usenet and Forums, respectively. 

\begin{figure*}[p]
    \centering
    \includegraphics[width=1\linewidth]{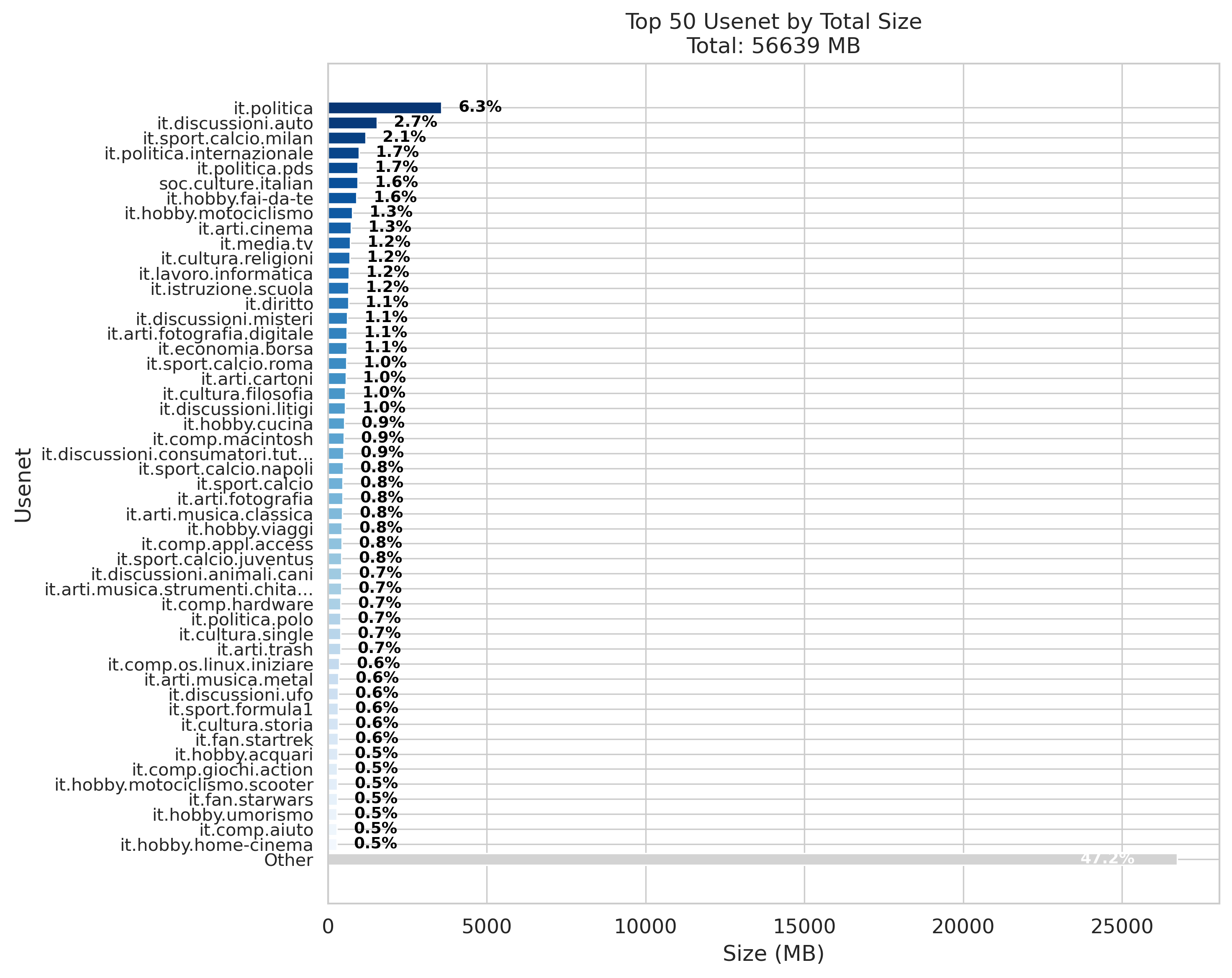}
    \caption{Top 50 newsgroups by total character count (all periods combined).}
    \label{fig:newsgroups_overall}
    \centering
    \includegraphics[width=1\linewidth]{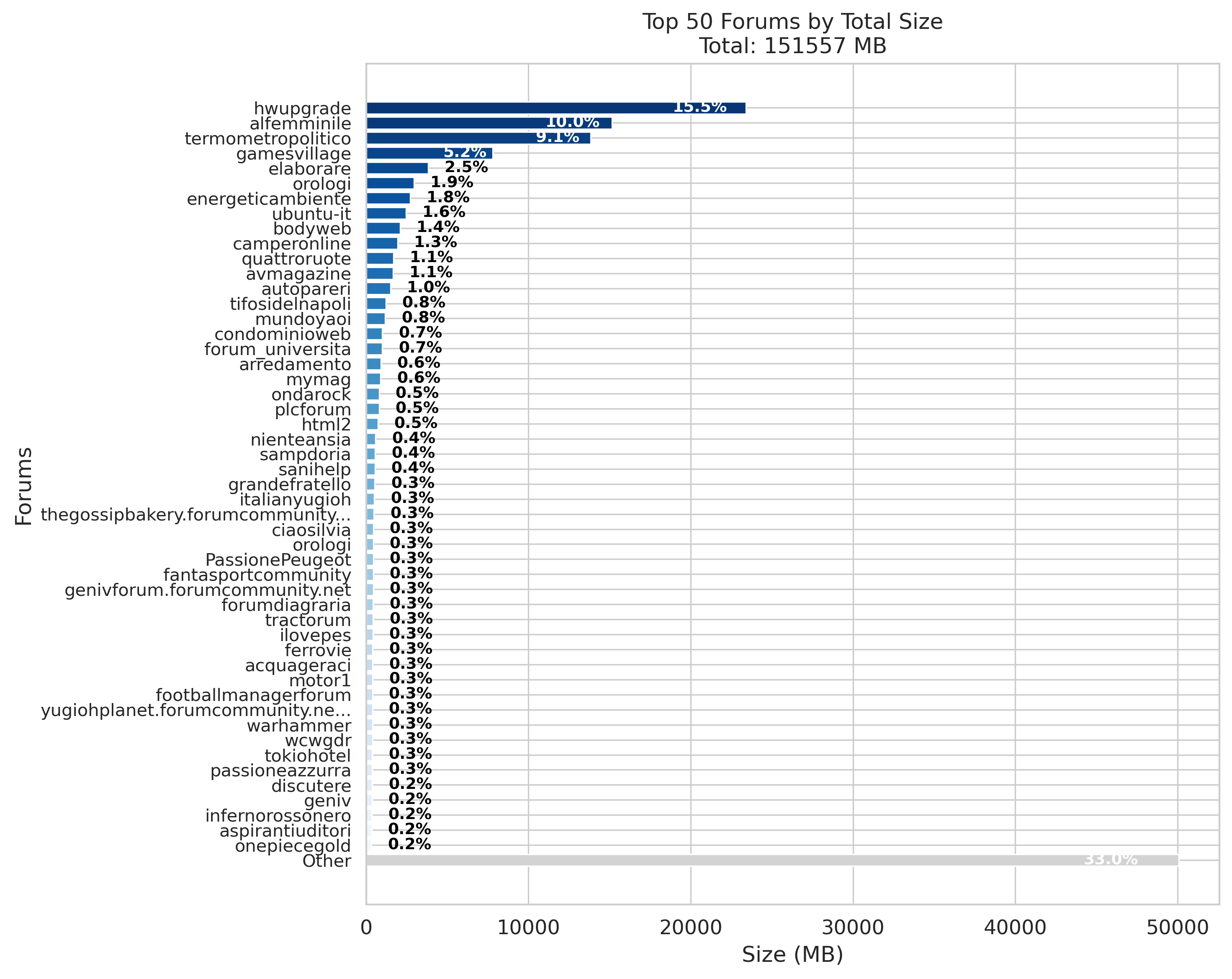}
    \caption{Top 50 forums by total character count (all periods combined).}
    \label{fig:forums_overall}
\end{figure*}

The topics covered by the posts are of different nature and can be inferred from the names of newsgroups (fig.\ref{fig:newsgroups_overall}) or forums (fig.\ref{fig:forums_overall}). Among the Usenet section, politics is the first topic covered by the resource, with \textit{it.politica} covering around 6\% of the data. Cars and soccer follow as more represented topics. Among forums, the first source is \textit{hwupgrade}, an Italian forum on technology, which represents around 15\% of the forum section. The second ranked, \textit{alfemminile}, is a forum devoted to women conversations, with topics that cover pregnancy, maternity, menstruation, among many others. As for Usenet, politics covers a significant proportion of the dataset, i.e. around 9\% of forums data.


\begin{figure*}[htbp]
    \centering
    \includegraphics[width=1\linewidth]{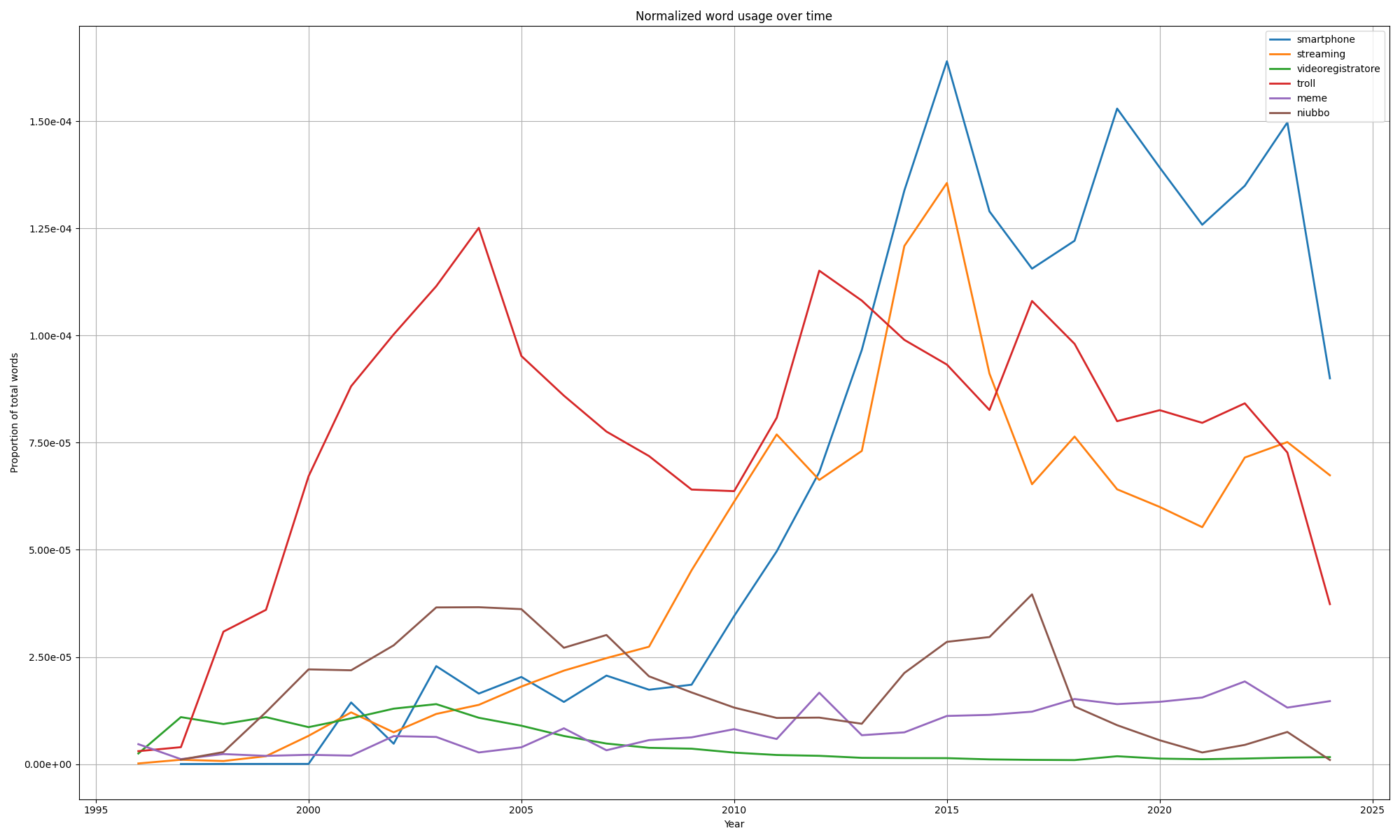}
    \caption{Normalized frequencies of six words across time in the \textsc{TestiMole-Conversational} corpus.}
    \label{fig:neologisms}
\end{figure*}

The corpus is released together with words frequency lists. Given the diachronic nature of the corpus, frequencies can be used in the socio-linguistic analysis to observe the rising and fall of specific terms in conversation or to identify neologisms. Figure \ref{fig:neologisms} shows the use of six words over the time period of the \textsc{TestiMole} corpus. It shows the rapid growth of use of the neologism \textit{troll}, which was coined during the first years of Usenet groups, and of the neologisms \textit{smartphone} and \textit{streaming}, which appears already on 2001 but gained popularity starting from 2010.

\section{The \textsc{TestiMole} Dataset}
\textsc{TestiMole-Conversational} is part of a larger dataset originally created in order to provide the academic community with better resources to train different kind of language models employing high-quality native Italian resources, also for long-context training. In this work, we decided to focus on the "conversational" subset as a scientific object on its own, given its relevance for a broad range of studies. It is worth to notice that alongside \textsc{TestiMole-Conversational}, the \textsc{TestiMole} dataset for the Italian language also comprises large amounts of cleaned textual data drawn from public domain books (2GB)\footnote{Currently, public domain books were mainly sourced from the LiberLiber's 'Progetto Manuzio' \url{https://liberliber.it/} } as well as open-access academic material (20GB), blogs (15GB) and the collection of several already existing Italian large corpora.

\section{Conclusion}
In an historical period characterized by the rise of Large Language Models and the consequent quest for large and clean datasets, \textsc{TestiMole} stands out as an important resource for improving the capabilities of natively Italian as well as multilingual LMs to correctly model peculiar elements of Italian language and society, drawing from the rich, diverse and collectively created shared knowledge base compiled by users during thirty years of Computer Mediated Interaction. This material can help models grasp conversational nuances typical of Italian discourse and guide them toward more natural problem-solving paths, potentially richer than those inferred solely from neutral sources. Furthermore, this conversational content can enhance emotion understanding capabilities, leading to improved classification systems for Italian data.
Given the proven importance of discussion board analysis for linguistic and sociological research, this resource offers an opportunity to introduce large-scale data analysis into such studies, which are often constrained by limited datasets, thereby reinforcing their experimental validity and unlocking possibilities for novel investigations. In this perspective, future work may include providing linguistic annotation on the  collected data, including for example lemmatization, POS tagging and syntactic parsing.

\section{Acknowledgements}
The work of V. Patti and M. Rinaldi have been partially supported by the “HARMONIA” project - M4-C2, I1.3 Partenariati Estesi - Cascade Call - FAIR - CUP C63C22000770006 - PE PE0000013 under the NextGenerationEU programme.

\section{Limitations}
Given the substantial manual work involved in designing appropriate collection strategies from diverse platforms, it was not possible to include every Italian discussion board in the corpus; neither it is possible to quantify the proportion of the collected resource over the total. Further sources could have probably been retrieved, but we believe that the present corpus already represents a wide and representative sample of this variety of CMC language.

It is also important to note that, although moderation was in place for many of the collected resources, we cannot guarantee that the corpus is free from profanity, offensive or aggressive language. Indeed, we did not aim to remove it, since the resource may be used for the study of hate speech as well, even from a diachronic point of view. However, while these registers may be of high interest for specific socio-linguistic research, their usage in language modelling should always be considered in relation to the intended use cases.

Finally, discussion board material may introduce unwanted noise in LLM training compared with data obtained from cleaner sources such as books, encyclopedias or academic papers: users may have introduced, either for amusement or actual misinformation purposes, erroneous or misleading information.

\section{Ethical considerations}
From an ethical standpoint, the collection of online conversational data raises concerns regarding user privacy and consent, even when such content was publicly accessible at the time of collection. To mitigate these risks, we anonymized all usernames from the corpus. We assume that users followed platform guidelines prohibiting the sharing of personal information; however, we acknowledge that inadvertent disclosure of sensitive details may still occur in user-generated content. Researchers using this dataset should be aware of these limitations and exercise caution when analysing or presenting findings that might compromise individual privacy.

According to the current version of the European Digital Services Act (Reg. UE 2022/2065), researchers can download publicly available data from web platforms for research purposes. Indeed, our resource is meant only for research, with no commercial intention. Accordingly, we plan to redistribute the data exclusively to researchers for research purposes, on a case-by-case basis, under the condition that applicants complete a request form including an end-user non-commercial license agreement.

\section{Appendix}
The script used to web-scrape the data is composed of four 
main functions:

\begin{itemize}
\item The \textbf{main} function, given an URL pattern defined by a prefix, a range, and a suffix, calls the \texttt{get\_url} function for all forum topics.
\item The \texttt{get\_url} function connects to the server and attempts to retrieve the page. In particular, it uses Python's \texttt{requests} library to download the URL, logging whether the download was successful or an error occurred. It then passes the downloaded HTML page to the BeautifulSoup parser in order to obtain the object required for subsequent parsing. Any BeautifulSoup errors are also logged.
\item The \texttt{extract\_thread} function, after having identified the discussion title, iterates through all the post HTML blocks (most often, \texttt{div} tags) and extracts the relevant information (author, datetime, with optional conversion to ISO format, and message body). If pagination is detected in the thread, the function is called iteratively (with the flag \texttt{inside\_pagination} set to \texttt{True}) for all the thread's pages until all the messages have been parsed. All this information is stored in a list of Python dictionaries: each element of the list corresponds to a post, while the list itself represents the thread. This list is then passed to the \texttt{save\_post\_to\_jsonl} function.
\item The \texttt{save\_post\_to\_jsonl} function iterates over the list and saves each message to a JSONL file, as a single row. Since Python lists are ordered collections, it also reconstructs the order of the posts, which is recorded in the JSONL entry under the \texttt{progressive\_number} column. An alternative data structure, such as a single JSONL entry per thread containing a list of all the posts, would have been more efficient; however, the less efficient approach was chosen to ensure better visualisation within the online Hugging Face platform. Conversion between these different storage formats is nevertheless trivial.
\end{itemize}

\section{Bibliographical References}\label{sec:reference}

\bibliographystyle{lrec2026-natbib}
\bibliography{lrec2026-example}

\section{Language Resource References}
\label{lr:ref}
\bibliographystylelanguageresource{lrec2026-natbib}
\bibliographylanguageresource{languageresource}

\end{document}